\documentclass[journal]{IEEEtran}
\usepackage[font=small,skip=1pt]{caption}
\IEEEoverridecommandlockouts
\usepackage{textcomp}
\usepackage{gensymb}
\usepackage{latexsym}
\usepackage[T1]{fontenc}
\usepackage{multirow}
\usepackage{graphicx}
\usepackage{varioref}
\usepackage{array}
\usepackage{enumitem}
\usepackage{soul,color}
\usepackage{longtable}
\usepackage{algorithmic}
\usepackage[ruled, vlined]{algorithm2e}
\usepackage{pifont}
\usepackage{booktabs}

\usepackage{color}
\definecolor{light}{rgb}{0.5, 0.5, 0.5}

\usepackage{graphicx}
\medskip
\usepackage{amssymb}
\usepackage{amsmath,epsfig,amssymb}
\usepackage{amsthm}
\usepackage{xcolor}
\usepackage[active]{srcltx} % SRC Specials for DVI Searching
\usepackage{epstopdf} % eps and png can work together

\usepackage{bm}
\usepackage{amsthm}
\usepackage{enumitem}
\usepackage{subfigure}
\DeclareRobustCommand*{\IEEEauthorrefmark}[1]{%
  \raisebox{0pt}[0pt][0pt]{\textsuperscript{\footnotesize\ensuremath{#1}}}}

\usepackage{cite}
\usepackage{amsmath,amssymb,amsfonts}
\usepackage{algorithmic}
\usepackage{graphicx}
\usepackage{textcomp}
\usepackage{xcolor}
\def\BibTeX{{\rm B\kern-.05em{\sc i\kern-.025em b}\kern-.08em
    T\kern-.1667em\lower.7ex\hbox{E}\kern-.125emX}}
\usepackage{lineno,hyperref}
\usepackage[utf8]{inputenc}
\usepackage[english]{babel}
\usepackage{mathdots}

\graphicspath{{Pictures/}{../jpeg/}} 
\usepackage{float}
\usepackage{amsmath}

\begin{document}

\title{Impact Learning: A Learning Method from Feature’s Impact and Competition}

%%%%%%

%\author[1]{Abu Jafar Md Muzahid}
%\author[2]{Syafiq Fauzi Kamarulzaman, \textit{Member, IEEE}}
%\author[3]{Md Arafatur Rahman, \textit{Senior Member, IEEE}}
%\author[4]{Md Abdus Samad Kamal, \textit{Senior Member, IEEE}}
%\affil[1]{Faculty of Computing, Universiti Malaysia Pahang, Malaysia.} % \authorcr Email: {\tt \{uid1, uid2\}@usc.edu}\vspace{1.5ex}}
%\affil[2]{Fellow of Automotive Engineering Center, University Malaysia Pahang, Malaysia.} %CA \authorcr Email: {\tt uid3@jpl.nasa.gov} \vspace{-2ex}}
%\affil[3]{School of Mathematics and Computer Science, University of Wolverhampton, UK}
%\affil[4]{Graduate School of Science and Technology, Gunma University, Japan.}

%%%%

\author{\IEEEauthorblockN{Nusrat Jahan Prottasha\IEEEauthorrefmark{1},
Saydul Akbar Murad\IEEEauthorrefmark{2},
Abu Jafar Md Muzahid\IEEEauthorrefmark{2}, 
Masud Rana\IEEEauthorrefmark{3},
Md Kowsher\IEEEauthorrefmark{4},
Apurba Adhikary\IEEEauthorrefmark{3},
Sujit Biswas\IEEEauthorrefmark{5} and
Anupam Kumar Bairagi\IEEEauthorrefmark{6}}\\
\vspace{1.5ex}
\IEEEauthorblockA{\IEEEauthorrefmark{1}Computer Science, Daffodil International University, Bangladesh.}\\
\IEEEauthorblockA{\IEEEauthorrefmark{2}Faculty of Computing, College of Computing \& Applied Sciences, Universiti Malaysia Pahang, Malaysia.}\\
\IEEEauthorblockA{\IEEEauthorrefmark{3}Information and Communication Engineering, Noakhali Science and Technology University, Bangladesh.}\\
\IEEEauthorblockA{\IEEEauthorrefmark{4}Computer Science, Stevens Institute of Technology, USA.}\\
\IEEEauthorblockA{\IEEEauthorrefmark{5}Computer Science, University of East London, UK.}\\
\IEEEauthorblockA{\IEEEauthorrefmark{6} Computer Science and Engineering Discipline, Khulna University, Bangladesh.}}  % <-this % stops an unwanted space

%\thanks{Manuscript received XXXXXXXX XX, XXXX; revised XXXXXXXX XX, XXXX. Corresponding author: Syafiq Fauzi Kamarulzaman (email:syafiq29@ump.edu.my)}

\markboth{Journal of \LaTeX\ Class Files,~Vol.~XX, No.~XX, XXXX~XXXX}{Shell \MakeLowercase{\textit{et al.}}: Bare Demo of IEEEtran.cls for IEEE Transactions on Magnetics Journals}

\maketitle
\begin{abstract}
Machine learning is the study of computer algorithms that can automatically improve based on data and experience. Machine learning algorithms build a model from sample data, called training data, to make predictions or judgments without being explicitly programmed to do so. A variety of well-known machine learning algorithms have been developed for use in the field of computer science to analyze data. This paper introduced a new machine learning algorithm called impact learning. Impact learning is a supervised learning algorithm that can be consolidated in both classification and regression problems. It can furthermore manifest its superiority in analyzing competitive data. This algorithm is remarkable for learning from the competitive situation and the competition comes from the effects of autonomous features. It is prepared by the impacts of the highlights from the intrinsic rate of natural increase (RNI). We, moreover, manifest the prevalence of the impact learning over the conventional machine learning algorithm.
\end{abstract} 
\begin{IEEEkeywords}
Impact learning, Machine Learning, Classification, Regression, Asthma prediction, Diabetes Prediction, Heart disease identification.
\end{IEEEkeywords}

\section{Introduction}
\label{sec:1}
Machine learning (ML) is a state-of-the-art approach that has shown promise in the areas of categorization and prediction. To improve demand estimates, we can use a variety of methods to examine historical data, including time series analysis, machine learning techniques, and deep learning models. As needed, ML ensures program consistency and adaptability. Machine learning, while not competitive, will continue to grow in the near future due to increasing data capital and greater need for personalized applications (e.g., the development of matrix multiplications) \cite{r1}. In addition to app growth, ML is also expected to change the general perspective of computer science. ML emphasizes creating a self-monitoring, self-diagnosing, and self-repairing system by shifting the focus from "how to program a machine" to "how to make it program itself." Both statistics and computer science can contribute to the evolution of ML as they develop and apply increasingly complex ideas that change the way people think \cite{r2}. In statistics and machine learning, extracting knowledge from data is an important endeavor. Many fields, including biomedicine, rely on it \cite{r3}, Business Analytics \cite{r4}, Computational Optimization \cite{r5}, Criminal Justice \cite{r6}, Cybersecurity \cite{r7}, Policy Making \cite{r8}, Process Monitoring, Regulatory Benchmarking\cite{r9}. Mathematical Optimization plays a vital role in building such models.

Classification and Regression Modern approaches based on recursive partitioning are trees. They are theoretically simple and exhibit exceptional learning performance. Nevertheless, they are very computationally expensive. There are methods and packages to instruct them in common programming languages such as Python and R. They are desirable not only in terms of their interpretability, but also because of their rule-based nature \cite{r10}. This makes it attractive for a variety of applications, such as credit scoring for lending. They used a dataset of individuals that includes demographic and financial predictors, among others, and the model uses this information to predict whether consumers will be good or bad payers.

%\textit{Figure \ref{stat}} present the World Health Organization's indication.

Each machine learning algorithm has its own application \cite{r11}. An approach may be optimal for a particular data set, but not for others. In real life, we work with directed data such as time periods (e.g., day, week, month, etc.), orientation, and rotation. Special algorithms are required to manage directed data. A small number of researchers, including \cite{r12}, have developed a non-probabilistic model for directional data. \cite{r12} have proposed a set of directional Support Vector Machines: updated SVM decision function. Using cosine and triangular waves, they study the periodic parametric mapping of directional variables. Moreover, they modified the model with triangular waves to allow asymmetric circular boundaries and kernelized the different SVM variants. Nevertheless, the additional factors involved in asymmetric SVMs periodically affect the decision boundary. We generate a large amount of directional data on a daily basis. However, the standard statistical distribution is inappropriate for this type of data. Therefore, different distributions and statistics should be used to study this type of data, such as the univariate von Mises distribution and the multivariate von Mises-Fisher distribution. A unidirectional Naive Bayes classifier predictor variable was demonstrated by \cite{r13}. Von Mises univariate or von Mises univariate must be used to apply this model. The predictor variables are modeled using the von Mises-Fisher and Gaussian distributions. The parameters of a Gaussian distribution have been found to affect the degree of complexity of the decision surfaces in a hybrid context. Von Mises Naive Bayes (vMNB) was tested on eight datasets and compared to the NB classifiers that use Gaussian distributions or discretization to model angular variables. The Selective NB classifier was, among others, the best algorithm for classifying. A generative probabilistic model, called latent Dirichlet assignment (LDA), was presented by \cite{r14}. They showed how to estimate empirical Bayes parameters using variational approaches and a EM algorithm, as well as other useful approximate inference techniques. Document modeling, collaborative filtering, and text classification were all applications of their findings. Tree Augmented Naive Bayes (TAN) outperforms naive Bayes, as shown by \cite{r15}. Nonetheless, the TAN maintained the same computational simplicity and resilience as the Bayes model. Similar work has been completed by \cite{r16} for categorizing Cell Anemia using a Deep neural network.

Different machine learning algorithms perform better for different types of data. In this manner, there are no "best" learning algorithms to be declared \cite{r17}. We are living in an age where the undertakings are information-driven, and the world is delivering more information than natural resources. There is a typical way to deal with while building up an algorithm, that is to plan a cost function and further minimize the cost function. Practically all learning algorithms attempt to limit a cost function for a superior and streamlined outcome. \cite{r18} presents an outline of a few of the strategies for testing linear separability between two classes. The techniques are separated into four gatherings: Those dependent on linear programming, those dependent on computational math, one dependent on neural networks, and one dependent on quadratic programming.\cite{r19} depict and examine a productive boosting system that can be utilized for limiting the loss functions got from our group of relaxations. \cite{r20} explores how concealed layers of profound rectifier networks are equipped for changing at least two example sets to be linearly distinguishable while safeguarding the distances with an ensured degree and demonstrates the all-inclusive characterization force of such distance saving rectifier organizations. \cite{r21} outline SVM utilizing a two-class issue and start with a case in which the classes are linearly divisible, implying that a straight line can be drawn that consummately isolates the classes, with the edge being the opposite distance between the nearest focuses to the line from each class.  \cite{r22} Murad uses recent developments in probabilistic machine learning to identify conditions arising from parametric linear operators. These conditions include, but are not limited to, the conventional and partial differential, fractional order operators, and  integrodifferential. \cite{r23} studies the utilization of logistic regression in manufacturing failures recognition. Using the generalized linear model for logistic regression makes it possible to break down the influence of the variables under study.

The core purpose of classification and regression is to predict the outcome from a single multidimensional data source. Regression attempts to fit a line with the data curve, whereas classification finds classes from the labeled data. The full functionality of an algorithm mimics a mathematical function. In \cite{r17}, we proposed a robust machine learning algorithm known as Impact Learning, where the main aim was to solve the classification and regression problems in machine learning applications. This article will briefly describe that algorithm for better understanding. If (\textit{X}) represents the data and (\textit{Y}) represents the outcome, then the model (F) would look like a function that converts \textit{X} into \textit{Y} [F(X)=Y]. This function can be called a mapping function.

In statistics and population studies, the rate of natural increase (RNI) is a well-known terminology \cite{r18}. ata is collected from real-world applications, and therefore, to keep it independent and distinguished, we can utilize (RNI) for generating a machine learning model. The following expression can represent RNI:

\begin{equation}\label{eq1}
\frac{dP}{dt} \approx  rP,
\end{equation}

where r is taken to be as RNI and P define as population size. To extend the equation by incorporating a logistic growth model \cite{r19} we find,

\begin{equation}\label{eq2}
\frac{dP}{dt} \approx  rP (1 - \frac{p}{k}),
\end{equation}

A dataset is comprised of numerous features. Almost every component of the dataset has RNI characteristics. In a competitive scenario, back forces are detrimental to the elements. Along these lines, the target variable gets affected by different highlights of the back forces, and we name that "Back Impact on Target (BIT)". Since the objective element depends on BITs, that is the reason each BIT additionally relies upon the objective factor.

In this paper, we intend to uncover the presence of impact learning.  Here we have utilized three kinds of datasets to verify the impact learning. We likewise showed the graphical and factual correlations among all current machine learning techniques such as Random Forest tree, Naive Bayes, SVM, Logistic regression, etc. The main contributions of this work are presented as follows:

\begin{itemize}
    \item We proposed an algorithm for supervised machine learning called impact learning that is used to utilize classification and regression datasets.
    \item We used RNI as a guide, this method makes use of the back impact of previous features.
    \item This algorithm is used to analyze the real competition as it learns from the competition.
    \item On the basis of a real dataset, we conducted a comparison between impact learning and the remaining algorithms.
\end{itemize}

\section{Introduction of Impact Learning}
\label{sec:2}

To visualize the exponential growth of population increment, we use the Malthusian logistic growth model\cite{new1}, which can be expressed as follow: 

\begin{equation}\label{eq3}
\frac{dy}{dt} \approx  ry,
\end{equation}

where “y” indicates the population and “r” is the RNI. By incorporating carrying capacity (K) with equation \ref{eq3} we get:

\begin{equation}\label{eq4}
\frac{dy}{dt} \approx  ry (1 - \frac{y}{k}).
\end{equation}

\begin{figure}[htb]
    \centering
    \includegraphics[width=0.8\linewidth]{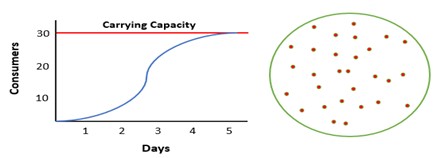}
    \caption{Whenever we have one operator(independent variable) only, the RNI den depends on only number of sample (consumers). The rate decreases with the time passing. }
    \vspace{-5pt}
    \label{fig1}
\end{figure}

Carrying capacity plays a crucial role in population dynamics. The population of a region cannot exceed a certain limit in ecology. Therefore, the growth rate in equation \ref{eq4}, has to be reversed after a particular time and gradually hit zero. We can visualize the phenomena through Figure \ref{fig1}.

\begin{figure}[htb]
    \centering
    \includegraphics[width=0.8\linewidth]{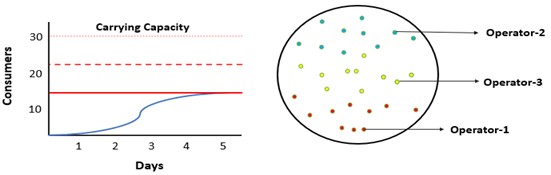}
    \caption{In thing figure, the RNI of operator-1 depends on other operator also and it is limited by its carrying capacity because of other operators.}
    \vspace{-5pt}
    \label{fig2}
\end{figure}

In Figure \ref{fig2}, we see a logistic growth model of cellphone purchases where the RNI curve increases exponentially from 0 to day 3, and after that, the rate decreases. We observe that the purchase impacts itself since a rise in the purchases automatically obstructs its growth at a certain point. Likewise, the presence of different operators can affect each other. If a particular operator makes profits, other operators face a decline simultaneously.

Back-impact on target prevents target features from moving toward the curve of the RNI (BIT). If x is the back-impact variable, then x and y maintain their effects on each other, but y maintains its effects on itself. So, we can rewrite the equation \ref{eq4} as

\begin{align*} 
\frac{dy}{dt}  & = ry - w_{x\longrightarrow y}xy -  w_{y\longrightarrow x}xy\\
              & = ry -  (w_{x\longrightarrow y} + w_{y\longrightarrow x})xy \\
             &  = ry - w * xy 
\end{align*}
\begin{equation}\label{eq5}
Or,  \frac{dy}{dt}  = ry - w_y y^2 - wxy,
\end{equation}

To visually explain,

\begin{figure}[htb]
    \centering
    \includegraphics[width=0.8\linewidth]{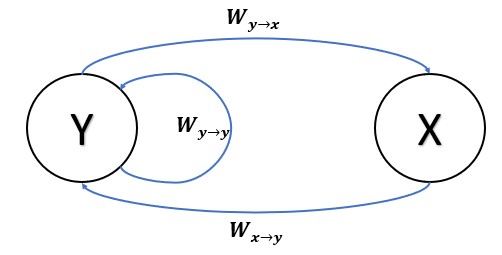}
    \caption{Here y is a target variable which has two dependence, one dependence is for other variable and another one is self dependence.  }
    \vspace{-5pt}
    \label{fig3}
\end{figure}

Now, combining the equation \ref{eq3}, \ref{eq4}, and \ref{eq5} we get

\begin{align*} 
ry (1-\frac{y}{x})  & = ry - w_y y^2 -  wxy\\
Or, r(1-\frac{y}{x}) & = r - w_y y -  wx \\
Or, y(\frac{r}{k} - w_y) & = -wx
\end{align*}

\begin{equation}\label{eq6}
Or, y = \frac{kwx}{r - w_y k} + b,
\end{equation}

To determine the impact of $x_k$ target feature y (trained) and $y^\prime$ is the target feature, then we get from the \ref{eq6} as follows:

\begin{equation}\label{eq7}
\operatorname{Imp}(\mathrm{y})=\left(y^{\prime}-\left(\frac{k \sum_{i=1}^{n} w_{i} x_{i}}{r-w_{y} k}+b\right)\right)^{\frac{2}{N}} \text { if } \mathrm{i} \neq \mathrm{k},
\end{equation}

where, N is the size of the dataset.

If X = [$x_1,x_2,x_3,.......,x_n$] and W = [$w_1,w_2,w_3,........,w_n$], then the \ref{eq6} can be expressed as follows.

\begin{equation}\label{eq8}
y = \frac{k(w^T .x)}{r - w_y k} + b.
\end{equation}

\begin{figure}[htb]
    \centering
    \includegraphics[width=0.8\linewidth]{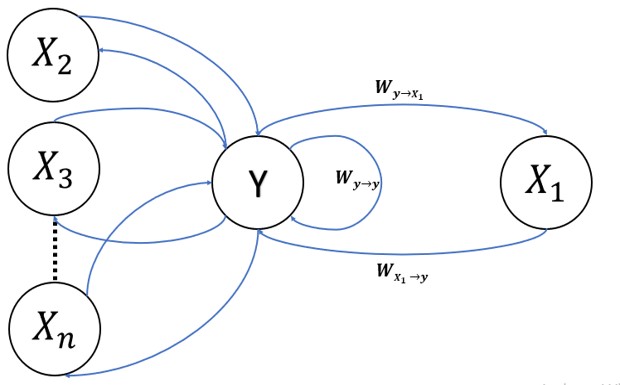}
    \caption{For multiple variable the constructed dependence.   }
    \vspace{-5pt}
    \label{fig4}
\end{figure}

\begin{figure}[htb]
    \centering
    \includegraphics[width=0.5\linewidth]{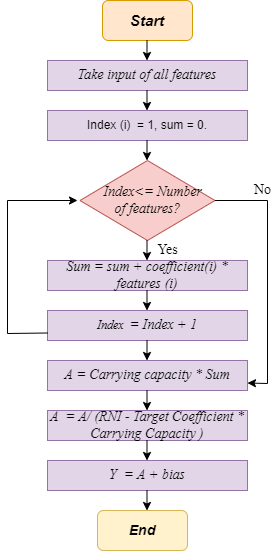}
    \caption{Workflow of calculating impact learning step by step}
    \vspace{-5pt}
    \label{fig5}
\end{figure}

In order to illustrate the polynomial structure of the impact learning, the \ref{eq6} can be further expressed as follows:

\begin{equation}\label{eq9}
 y=\frac{k \sum_{i=1}^{n} w_{i} x_{i}^{j}}{r-w_{y} k}+b,
\end{equation}

where j>0.
For the N size of data, the RNI and carrying capacity can be found from these defined functions.

\begin{equation}\label{eq10}
r=\frac{\ln \frac{\max \left(y^{\prime}\right)}{\min \left(y^{\prime}\right)}}{N-1} \text { and } k \geq \max \left(y^{\prime}\right)
\end{equation}

However, it is an excellent way to calculate the RNI (r) from the optimization techniques like gradient descent. Figure \ref{fig5} describes the flow chat of the proposed algorithm which is mathematically expressed in \ref{eq1} above.

\section{Experimental Results and Discussion}
\label{sec:3}

We used the proposed method to tackle various real-world situations, demonstrating the acceptability of the Impact Learning model. In addition, we compared the performance of the proposed approach to that of other standard machine learning techniques (Random Forest, KNN, SVM, Naïve Bayes, Linear Regression and Logistic Regression). Besides, we have employed a variety of datasets. The dataset, training and testing procedure, experimental method, outcome, and commentary will all be covered in the following sub-sections.

\subsection{Datasets}
All of the datasets we have used to verify the proposed approach have been gathered from the real-world environment. We constructed a Google form for each dataset to collect the information. Our first dataset is for the prediction of diabetes. We created two Google forms to collect data from participants by asking them various questions. A separate form exists for patients and another for non-patients; however, all fields are identical in both formats. The total data obtained is 469, with 232 non-diabetic patients and 237 diabetic patients included. There are 11 attributes of the first dataset, shown in Table \ref{tab1}.

% \usepackage{booktabs}
\begin{table*}[]
\centering
\caption{Dataset attributes for diabetes identification.\label{tab1}}
\begin{tabular}{@{}lll@{}}
\toprule
\textbf{Attributes}    & \textbf{Description}                                                                                             & \textbf{Type} \\ \midrule
Age                    & Age of patient.                                                                                                  & Numeral       \\
Weight                 & Weight of the patient.                                                                                           & Numeral       \\
Family History         & \begin{tabular}[c]{@{}l@{}}If the family member has diabetes history, \\ than ‘Yes’ otherwise ‘No’.\end{tabular} & Boolean       \\
Late-night sleep habit & It’s defined the sleeping habit of patient.                                                                      & Boolean       \\
Heart Diseases         & If the patient has heart disease, then ‘Yes’ otherwise ‘No’.                                                     & Boolean       \\
Sleep after eating     & If the patient goes to bed after eating, then ‘Yes’ otherwise ‘No’.                                              & Boolean       \\
Addiction              & This field want to know about drug addiction of patient.                                                         & Boolean       \\
Sex                    & Gender of patient.                                                                                               & Boolean       \\
Late wake-up habit     & Wake-up habit of patient.                                                                                        & Boolean       \\
Exercise               & Exercise habit.                                                                                                  & Boolean       \\ \bottomrule
\end{tabular}
\end{table*}

Before collecting data for our second dataset on asthma, we consult with asthma professionals. We had taken a brief note from the doctor's description of the traits that cause asthma. Finally, we extracted 23 characteristics from them. Based on such features, we created a Google Form. We collected information through social media and through visits to the doctor's office. This Google form was distributed to a public group, shared on our timeline, and emailed to our contacts to collect data. We spent several days physically collecting data from asthma patients and those who exhibit asthmatic symptoms in the doctor's chamber. While some previous research collected data exclusively from youngsters, we obtained data from individuals of various ages in our study. We do not have any restrictions on age. Finally, we gathered almost 500 data points from social media and the chamber of visiting doctors. Table \ref{tab2} summarizes all  of the characteristics of the second dataset.

% \usepackage{booktabs}
\begin{table*}[]
\centering
\caption{Dataset attributes for asthma identification.}\label{tab2}
\begin{tabular}{@{}lll@{}}
\toprule
\textbf{Attributes}                                                                                                                                                    & \textbf{Description}                                                                                                                  & \textbf{Type} \\ \midrule
Gender                                                                                                                                                                 & If gender male then ‘M’ or 'F' for Female                                                                                          & Boolean       \\
Age                                                                                                                                                                    & Age of the patient                                                                                                                    & Numeral       \\
Frequent Coughing                                                                                                                                                      & \begin{tabular}[c]{@{}l@{}}It's define the continues coughing. If it's continues \\ then ‘YES otherwise ‘NO’\end{tabular}                     & Boolean       \\
Shortness of breath                                                                                                                                                    & It's define Breathing problems                                                                                         & Boolean       \\
\begin{tabular}[c]{@{}l@{}}Exercise or walks make \\ feel very tired\end{tabular}                                                                                      & If the person is tired, then ‘YES’ otherwise ‘NO’                                                                                     & Boolean       \\
\begin{tabular}[c]{@{}l@{}}Nausea or Coughing after\\  walking or exercise \end{tabular}                                                                                & \begin{tabular}[c]{@{}l@{}}Walking or Running for a while if he/she is coughing then ‘YES’ \\ otherwise ‘NO’\end{tabular}               & Boolean       \\
Sleep problems                                                                                                                                                         & If face difficulties in sleep then ‘YES’, otherwise ‘NO’                                                                              & Boolean       \\
\begin{tabular}[c]{@{}l@{}}There is a feeling of being\\  stuck in the chest\end{tabular}                                                                              & Usually, if you feel chest pain, answer 'YES'; otherwise, answer 'NO'                                                                         & Boolean       \\
Is anyone in family, suffers from allergies?                                                                                                                              & \begin{tabular}[c]{@{}l@{}}It’s define family history. If anyone has allergies, \\ then ‘YES’ otherwise ‘NO’\end{tabular}                   & Boolean       \\
Does any family member has asthma?                                                                                                                                 & \begin{tabular}[c]{@{}l@{}}It’s define family history. If any family has asthma, \\ then ‘YES’ otherwise ‘NO’\end{tabular} & Boolean       \\
Do you have allergies?                                                                                                                                                 & If have allergies, then ‘YES’ otherwise ‘NO’                                                                                          & Boolean       \\
\begin{tabular}[c]{@{}l@{}} Which season does shortness \\ of breath increase?\end{tabular}                                                                     & \begin{tabular}[c]{@{}l@{}}Here have four seasons. From there, Have to select one season, \\ when problem is increase.\end{tabular}                & Text          \\
\begin{tabular}[c]{@{}l@{}}Problem in Cold Water, Curd or Ice-cream, \\ beef, ilish, prawn, brinjal, pumpkin, coconuts, \\ Malabar night shade, duck meat\end{tabular} & \begin{tabular}[c]{@{}l@{}}If the patient feels this kind of disease then \\ ‘Yes’ otherwise ‘No’\end{tabular}                        & Boolean       \\
Have asthma?                                                                                                                                                           & If you have asthma put ‘YES’, otherwise ‘NO’                                                                                          & Boolean       \\ \bottomrule
\end{tabular}
\end{table*}

Our third dataset is utilized to forecast a heart disease prediction. This dataset was gathered from Kaggle and included a total of fifteen features \cite{new2}. The first fourteen variables (male, currentSmoker, age, BPMeds, cigsPerDay, prevalentStroke, prevalentHyp, totChol, diabetes,  sysBP, BMI, diaBP, glucose heartrate, , and region) are utilized as feature data, while the last variable (TenYearCHD) is used as level data. All of the implemented machine learning models performed well on this dataset.Table \ref{tab3} contains the attributes and description of the heart disease dataset.

% \usepackage{booktabs}
\begin{table*}[]
\centering
\caption{Dataset attributes for heart disease identification.\label{tab3}}
\begin{tabular}{@{}lll@{}}
\toprule
\textbf{Attributes} & \textbf{Description}                                                  & \textbf{Type} \\ \midrule
male                & It’s defined the sex: male or female.                                 & Nominal       \\
Age                & Patient age.                                 & Number       \\
currentSmoker       & whether the patient is a current smoker or not                        & Nominal        \\
cigsPerDay          & the average number of cigarettes smoked each day by the individual & Number        \\
BPMeds              & patient's blood pressure medicine           & Nominal       \\
prevalentStroke     & whether the patient had already experienced a stroke                & Nominal       \\
prevalentHyp        & whether the patient was hypertensive or not                           & Nominal       \\
diabetes            & Diabetes history of the patient                              & Nominal       \\
totChol             & It's define cholesterol level                                               & Number        \\
sysBP               & systolic blood pressure of patient                                               & Number        \\
diaBP               & Current diastolic blood pressure                                              & Number        \\
BMI                 & Body Mass Index of patient                                                      & Number        \\
heartRate           & heart rate of patient                                                           & Number        \\
glucose             & glucose level                                                         & Number        \\
TenYearCHD          & Coronary heart disease is a 10-year risk                            & Boolean       \\ \bottomrule
\end{tabular}
\end{table*}

\subsection{Experimental Setup}

\begin{figure}[htb]
    \centering
    \includegraphics[width=1\linewidth]{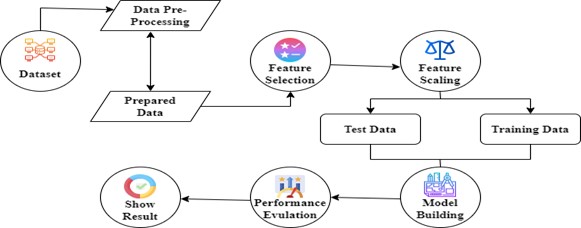}
    \caption{Workflow of our experimental task.}
    \vspace{-5pt}
    \label{fig6}
\end{figure}

We use a simulation environment to present the result in this paper. First, we devised a mathematical model. To implement this model, we wrote a Python script that ran on Google Colaboratory, a cloud computing platform with a Python programming environment. For quicker and parallel computing, we employed a Google Colab-based GPU. The development of all the models took more than 200 hours in total. We've also made a python module of this model available for public usage, which will make it much easier for consumers to use it. The proposed model was implemented in the following steps, as illustrated in Figure \ref{fig6}.

\subsubsection{Data Pre-processing and preparing data}

During data preprocessing, we determine whether a data set contains null values. After the one-time data processing, we checked if the data is prepared for the following process. If we determined that the data is ready for the next process, then we stop it \cite{r24}. We repeated it numerous times. We utilized mean and median to preprocess the data.A computation is used to determine the mean, synonymous with the average value of a data collection.

\subsubsection{Feature Selection}
\begin{figure}[htb]
    \centering
    \includegraphics[width=0.8\linewidth]{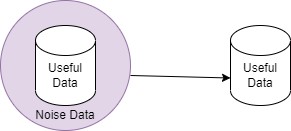}
    \caption{Feature Selection.}
    \vspace{-5pt}
    \label{fig7}
\end{figure}

Feature selection can be used to reduce the input variable model's size by selecting only relevant data and eliminating noise in the data. It is the process of automatically selecting suitable characteristics for the machine learning model utilized based on the type of problem being addressed. It is accomplished by incorporating or removing critical elements without altering their functionality. It aids in reducing noise in our data and the reduction of the size of our input data. Figure \ref{fig7} illustrates the diagram of feature selection. Feature selection models can be classified into supervised and unsupervised \cite{r25,r26}. We use a supervised learning model to analyze the data of our datasets. The term "supervised feature selection" refers to selecting features that make use of the output label class. They use the target variables to uncover variables that can improve model's efficiency and then incorporate these variables into the model.

\subsubsection{Feature Scaling}
In data analysis, feature scaling is a technique used to standardize the range of independent variables or characteristics of data \cite{r27,new3}. This happens in the preprocessing phase of data processing and is also called normalisation of data. Normalization and standardization are two of the most commonly used strategies for feature scaling. Normalization is utilized when we wish to confine our values to a range between two numbers, often between [0,1] and [-1,1], and we want to do so as efficiently as possible. The following is the general normalization formula:

\begin{equation}\label{eq11}
z^{\prime}=\frac{z-\min (z)}{\max (z)-\min (z).}
\end{equation}

Feature standardization reduces the data's variance and means to zero for each of the data's features. For each feature, the distribution mean and standard deviation are calculated, then the new data point is calculated using the formula below:

\begin{equation}\label{eq12}
z^{\prime}=\frac{z-\bar{z}}{\sigma}
\end{equation}

We separate our dataset into two parts in the feature scaling process: training and testing, with 70 percent of the data being used for training and 30 percent for testing, respectively. For the first and second datasets, we use Normalization; however, for datasets three and four, we use standardization.

\subsubsection{Model Building}
We employ four classification techniques ( Random Forest, k-nearest neighbors, support vector machine, and Naive Bayes) and two regression models (Linear regression and Logistic Regression) to utilize the entire dataset, except the newly proposed model. We compared our proposed model to the techniques employed to analyze the whole dataset. For our proposed model, we got a good outcome. A summary of all algorithms used in this work and their parameters are represented in Table \ref{tab4}.

\begin{table*}[]
\centering
\caption{This table contain the details of all implemented algorithms.\label{tab4}}
\begin{tabular}{lll}
\hline
\textbf{Name of Algorithm}   & \textbf{Description}                                                                                                                                               & \textbf{Parameters}                                                           \\ \hline
Random Forest                & \begin{tabular}[c]{@{}l@{}}Creates decision trees from various samples and uses the majority \\ vote for classification.\end{tabular}                              & \begin{tabular}[c]{@{}l@{}}Number of \\ decision tress\end{tabular}           \\
K-Nearest Neighbors (KNN)    & \begin{tabular}[c]{@{}l@{}}The new data point is classified by the majority of votes from its \\ fixed neighbors.\end{tabular}                                     & Nearest Neighbors.                                                            \\
Support Vector Machine (SVM) & \begin{tabular}[c]{@{}l@{}}Find a hyperplane in an N-dimensional space that clearly \\ classifies the data points (N — the number of characteristics)\end{tabular} & Kernel function.                                                              \\
Naïve Bayes                  & \begin{tabular}[c]{@{}l@{}}Classification method based on Bayes' Theorem and \\ the premise of predictor independence\end{tabular}                                 & \begin{tabular}[c]{@{}l@{}}Probabilities of \\ different classes\end{tabular} \\
Linear Regression            & \begin{tabular}[c]{@{}l@{}}Undertakes the task of predicting a dependent variable (target) \\ based on the independent variable provided (s).\end{tabular}         & \begin{tabular}[c]{@{}l@{}}Value of independent \\ variable\end{tabular}      \\
Logistic Regression          & \begin{tabular}[c]{@{}l@{}}It's the method of choice for binary classification issues \\ (problems with two class values)\end{tabular}                             & \begin{tabular}[c]{@{}l@{}}Value of all \\ independent variable\end{tabular}  \\ \hline
\end{tabular}
\end{table*}

\subsection{Performance Analysis}
This section is mainly divided into two sub-sections: performance analysis for classification algorithms and regression algorithms. The accuracy, precision, F1-score, recall, learning curves, and ROC curves for classification algorithms are used to examine the results of the experiments. These evaluation matrices are compared for four classification algorithms and two regression algorithms.

\subsubsection{Performance Analysis for Classification Algorithms (Asthma Dataset)}
We collected 500 data points from the real world to evaluate the asthma dataset, 211 from asthma patients and 289 from non-asthmatic patients. As a result of the fact that this is a balanced dataset, the accuracy score will provide an right idea for the prediction result. Furthermore, the evaluation should include accuracy and recall scores, and the F1 score provides a single harmonic mean score that accounts for both recall and precision. Our experiment used four robust machine learning models (Random Forest, Support Vector Machine, k-Nearest Neighbors, and Naive Bayes) and compared those to our proposed algorithm. The accuracy of the employed models is used to evaluate their performance.

\begin{figure}[htb]
    \centering
    \includegraphics[width=0.8\linewidth]{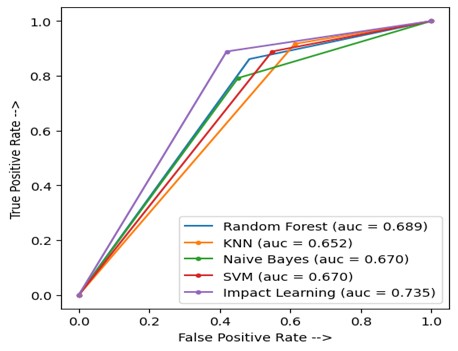}
    \caption{Comparison of all implemented algorithms with Impact Learning using ROC curve for asthma dataset.}
    \vspace{-5pt}
    \label{fig8}
\end{figure}

\begin{table}[]
\caption{Accuracy, Precision, F1-score and  Recall for all implemented classification algorithms and ranked them into ascending order. This performance metrics are for asthma Dataset.\label{tab5}}
\begin{tabular}{|llllll|}
\hline
\multicolumn{1}{|l|}{\textbf{\begin{tabular}[c]{@{}l@{}}Method \\ Name\end{tabular}}} & \multicolumn{1}{l|}{\textbf{AC}} & \multicolumn{1}{l|}{\textbf{PR}} & \multicolumn{1}{l|}{\textbf{RC}} & \multicolumn{1}{l|}{\textbf{\begin{tabular}[c]{@{}l@{}}F1-\\ Score\end{tabular}}} & \textbf{\begin{tabular}[c]{@{}l@{}}Rank \\ (Accuracy)\end{tabular}} \\ \hline
\multicolumn{1}{|l|}{\begin{tabular}[c]{@{}l@{}}Impact \\ learning\end{tabular}}      & \multicolumn{1}{l|}{0.806}       & \multicolumn{1}{l|}{0.761}       & \multicolumn{1}{l|}{0.734}       & \multicolumn{1}{l|}{0.745}                                                        & 1                                                                   \\ \hline
\multicolumn{1}{|l|}{SVM}                                                             & \multicolumn{1}{l|}{0.775}       & \multicolumn{1}{l|}{0.698}       & \multicolumn{1}{l|}{0.681}       & \multicolumn{1}{l|}{0.721}                                                        & 2                                                                   \\ \hline
\multicolumn{1}{|l|}{KNN}                                                             & \multicolumn{1}{l|}{0.757}       & \multicolumn{1}{l|}{0.721}       & \multicolumn{1}{l|}{0.653}       & \multicolumn{1}{l|}{0.661}                                                        & 3                                                                   \\ \hline
\multicolumn{1}{|l|}{\begin{tabular}[c]{@{}l@{}}Random \\ Forest\end{tabular}}        & \multicolumn{1}{l|}{0.752}       & \multicolumn{1}{l|}{0.683}       & \multicolumn{1}{l|}{0.711}       & \multicolumn{1}{l|}{0.691}                                                        & 4                                                                   \\ \hline
\multicolumn{1}{|l|}{\begin{tabular}[c]{@{}l@{}}Bernoulli-\\ NB\end{tabular}}         & \multicolumn{1}{l|}{0.714}       & \multicolumn{1}{l|}{0.665}       & \multicolumn{1}{l|}{0.671}       & \multicolumn{1}{l|}{0.662}                                                        & 5                                                                   \\ \hline
\multicolumn{6}{|l|}{AC = Accuracy, PR = Precision, RC = Recall}                                                                                                                                                                                                                                                                                         \\ \hline
\end{tabular}
\end{table}

From Table \ref{tab5}, with an accuracy of 80.6\%, the proposed algorithm (Impact learning) shows the best result compared to all other proposed algorithms. For the SVM, the accuracy is also good, which is 77.5\%. The KNN achieved an accuracy of 75.7\%, which is very similar to Random Forest. The lowest accuracy is found for BernoulliNB, which is 71.4\%. Following a comparison of the four types of performance matrices for various methods, it is clear that our suggested models (Impact Learning) perform the best in these datasets.

The receiver operating characteristic (ROC) plot is one of the most often used methods for assessing classifier performance. The ROC plot is built on two key evaluation metrics: specificity and sensitivity. The negative part's performance is measured by specificity, whereas the positive part's performance is measured by sensitivity\cite{r28}.

In the ideal condition, FPR = 0, indicating that the area under the ROC curve (AUC) equals 1, but in sub-optimal situations, it is less than 1. In Figure \ref{fig8}, the ROC curves for the asthma dataset are shown. The higher the curve, the better the results. The area under the ROC curve (AUC) for each classifier has been determined as a quantitative measure of performance. All of the methods we tested generated satisfactory results. With a score of 73.5 percent, Impact learning is better than other algorithms.

\begin{figure*}
    \centering
    \subfigure[]{\includegraphics[width=.42\textwidth]{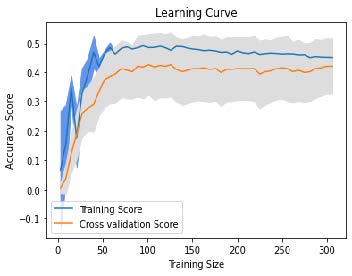}} 
    \subfigure[]{\includegraphics[width=.42\textwidth]{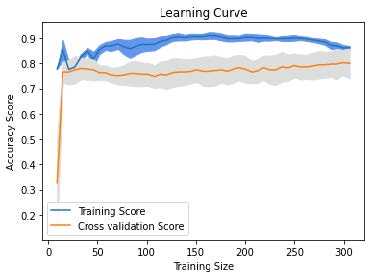}} 
    \subfigure[]{\includegraphics[width=.42\textwidth]{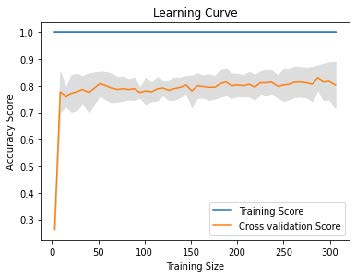}} 
    \subfigure[]{\includegraphics[width=.42\textwidth]{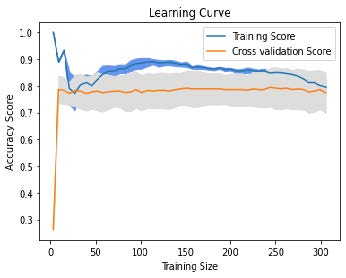}}
    \caption{Learning curve for the SVM, KNN, Random Forest and BernoulliNb classification algorithms which is used for asthma dataset. (a) Learning curve for SVM classifier. (b) Learning curve for KNN classifier. (c) Learning curve for Random Forest classifier. (d) Learning curve for BernoulliNB classifier. }
    \label{fig9}
\end{figure*}

The Learning Curve is a graphical tool that may be used to estimate the benefit of supplementing our model with extra training data. It demonstrates the relationship between training and test scores for a machine learning model with variable sample size. Generally, the cross-validation strategy is utilized when drawing the learning curve. We visualized the learning curve using the Python Yellowbrick tool. The word "Training Score" refers to the accuracy score for the train set, whereas the term "Cross-Validation Score" refers to the accuracy score for the test set.

\begin{figure}[htb]
    \centering
    \includegraphics[width=0.8\linewidth]{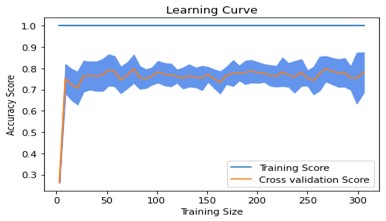}
    \caption{Learning curve for the proposed algorithm (Impact learning) that is used for asthma dataset.}
    \vspace{-5pt}
    \label{fig10}
\end{figure}

\begin{figure}[htb]
    \centering
    \includegraphics[width=0.8\linewidth]{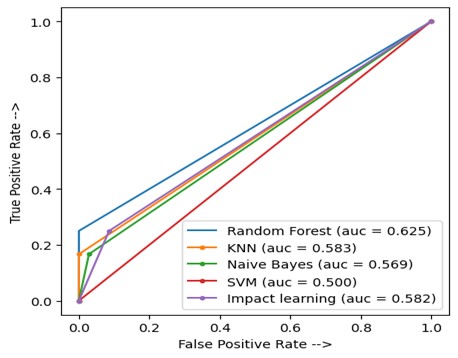}
    \caption{Comparison of all implemented algorithms with Impact Learning using ROC curve for diabetes dataset.}
    \vspace{-5pt}
    \label{fig11}
\end{figure}

\begin{figure*}
    \centering
    \subfigure[]{\includegraphics[width=.42\textwidth]{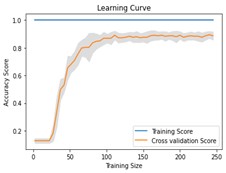}} 
    \subfigure[]{\includegraphics[width=.42\textwidth]{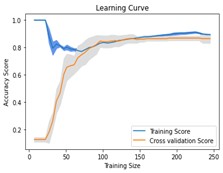}} 
    \subfigure[]{\includegraphics[width=.42\textwidth]{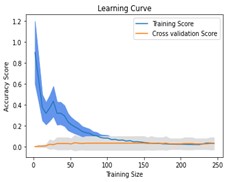}} 
    \subfigure[]{\includegraphics[width=.42\textwidth]{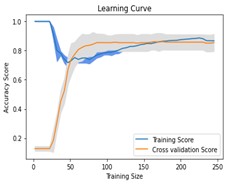}}
    \caption{Learning curve for the SVM, KNN, Random Forest and BernoulliNb classification algorithms which is used for asthma dataset. (a) Learning curve for Random Forest classifier. (b) Learning curve for KNN classifier. (c) Learning curve for SVM. (d) Learning curve for BernoulliNB classifier. }
    \label{fig12}
\end{figure*}

\begin{figure}[htb]
    \centering
    \includegraphics[width=0.8\linewidth]{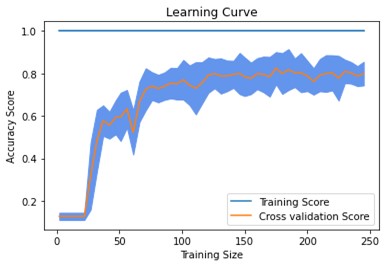}
    \caption{Learning curve for the proposed algorithm (Impact learning) that is used for diabetes dataset.}
    \vspace{-5pt}
    \label{fig13}
\end{figure}

Figure \ref{fig9} shows the learning curves of the four classification algorithms in terms of accuracy and speed. Figure \ref{fig9}(a) depicts the learning curve for the SVM classifier, which demonstrates that increasing the number of training cases leads to better generalization. This model's training and test scores are highly well-matched to the dataset. Figure \ref{fig9}(b) of the KNN classifier indicates that the training and test scores have not merged, showing that this model may benefit from more training. The Random Forest classifier (Figure \ref{fig9}(c)) indicates a significant degree of systematic error, which suggests that the model, regardless of the quantity of data fed into it, is incapable of reflecting the underlying relationship and has a high level of systematic error. With an increase in the training dataset, BernoulliNB (Figure \ref{fig9}(d)) shows that the model's performance will continue to improve. Impact learning (Figure \ref{fig10}) illustrates that the training and testing are still not working effectively. We need to enhance the dataset's size and diversity to get better results.

\subsubsection{Performance Analysis for Classification Algorithms (Diabetes Dataset)}
For the study of diabetes evolution, we have gathered data from the actual world, both physically and through an online platform. We have collected 469 data points, with 232 being non-diabetic and 237 being diabetic. Because the prevalence of diabetes and non-diabetes is nearly equal, this dataset is considered a balanced dataset. Accuracy is the most vital metric to consider while identifying the appropriate method for the balanced  dataset. This dataset’s accuracy, F1 score, precision, and recall have all evolved. We compare our proposed approach with four robust machine learning algorithms, shown in Table \ref{tab6}. According to the accuracy of the algorithms, we ranked them in Table \ref{tab6}.

\begin{table}[]
\caption{Accuracy, Precision, Recall and F1-score for all implemented classification algorithms and ranked them into ascending order. This performance metrics are for Diabetes Dataset.\label{tab6}}
\begin{tabular}{|llllll|}
\hline
\multicolumn{1}{|l|}{\textbf{\begin{tabular}[c]{@{}l@{}}Method \\ Name\end{tabular}}} & \multicolumn{1}{l|}{\textbf{AC}} & \multicolumn{1}{l|}{\textbf{PR}} & \multicolumn{1}{l|}{\textbf{RC}} & \multicolumn{1}{l|}{\textbf{\begin{tabular}[c]{@{}l@{}}F1-\\ Score\end{tabular}}} & \textbf{\begin{tabular}[c]{@{}l@{}}Rank \\ (Accuracy)\end{tabular}} \\ \hline
\multicolumn{1}{|l|}{\begin{tabular}[c]{@{}l@{}}Random \\ Forest\end{tabular}}        & \multicolumn{1}{l|}{0.89}        & \multicolumn{1}{l|}{0.94}        & \multicolumn{1}{l|}{0.62}        & \multicolumn{1}{l|}{0.89}                                                         & 1.00                                                                \\ \hline
\multicolumn{1}{|l|}{KNN}                                                             & \multicolumn{1}{l|}{0.87}        & \multicolumn{1}{l|}{0.93}        & \multicolumn{1}{l|}{0.58}        & \multicolumn{1}{l|}{0.60}                                                         & 2.00                                                                \\ \hline
\multicolumn{1}{|l|}{\begin{tabular}[c]{@{}l@{}}Impact \\ Learning\end{tabular}}      & \multicolumn{1}{l|}{0.86}        & \multicolumn{1}{l|}{0.60}        & \multicolumn{1}{l|}{0.58}        & \multicolumn{1}{l|}{0.59}                                                         & 3.00                                                                \\ \hline
\multicolumn{1}{|l|}{\begin{tabular}[c]{@{}l@{}}Bernoulli-\\ NB\end{tabular}}         & \multicolumn{1}{l|}{0.85}        & \multicolumn{1}{l|}{0.68}        & \multicolumn{1}{l|}{0.56}        & \multicolumn{1}{l|}{0.58}                                                         & 4.00                                                                \\ \hline
\multicolumn{1}{|l|}{SVM}                                                             & \multicolumn{1}{l|}{0.85}        & \multicolumn{1}{l|}{0.42}        & \multicolumn{1}{l|}{0.50}        & \multicolumn{1}{l|}{0.46}                                                         & 5.00                                                                \\ \hline
\multicolumn{6}{|l|}{AC = Accuracy, PR = Precision, RC = Recall}                                                                                                                                                                                                                                                                                         \\ \hline
\end{tabular}
\end{table}

Table \ref{tab6} shows that we found a good accuracy for all of the implemented algorithms. Compared to other algorithms, the random forest classifier's performance is superior because of its accuracy of 89 percent. The accuracy of the KNN classifier is 87.8 percent, which is the second-highest accuracy among the classifiers. The accuracy rate for Impact learning is 86 percent, which is likewise a strong performance rating. The BernoulliNB achieves an accuracy of 85.4 percent, nearly identical to that of the SVM. For all other (Precision, Recall, F1-Score) parameters, the performance is also good for the Impact learning.

Figure \ref{fig11} depicts the asthma dataset's receiver operating characteristic (ROC) curves. The higher the slope of the curve, the better the outcome. Each classifier’s area under the receiver operating characteristic curve (AUC) has been calculated as a quantitative measure of performance. For the majority of the algorithms, we were able to achieve a minimum satisfactory outcome. Compared to other algorithms, Random Forest performs significantly better, receiving a score of 62.5 percent. This score is 58.5 percent for impact learning, comparable to 58.5 percent for KNN. SVM received a low score, indicating that it was not the best choice for this dataset.

No matter how much data is fed into it, the Random Forest classifier (Figure \ref{fig12}(a)) cannot capture the underlying link accurately and has a significant systematic error. Figure 8 displays the learning curves of the four classification algorithms without the impact of learning. Underfitting can be seen by looking at the KNN's training score (Figure \ref{fig12}(b)), which continues to fall. In contrast, the cross-validation score (orange line) continues to rise and fall. based on the SVM's findings, Figure \ref{fig12}(c) illustrates how increasing the number of training instances improves generalization. Finally, the BurnolliNB (Figure \ref{fig12}(d)), increasing the number of training cases enhances generalization. Figure \ref{fig13} illustrates that the training and test scores for the Impact learning model have not yet converged, which suggests that this model could benefit from more training data being added.

\subsubsection{Performance Analysis for regression Algorithms (Heart disease Dataset)}

\begin{figure}[htb]
    \centering
    \includegraphics[width=0.8\linewidth]{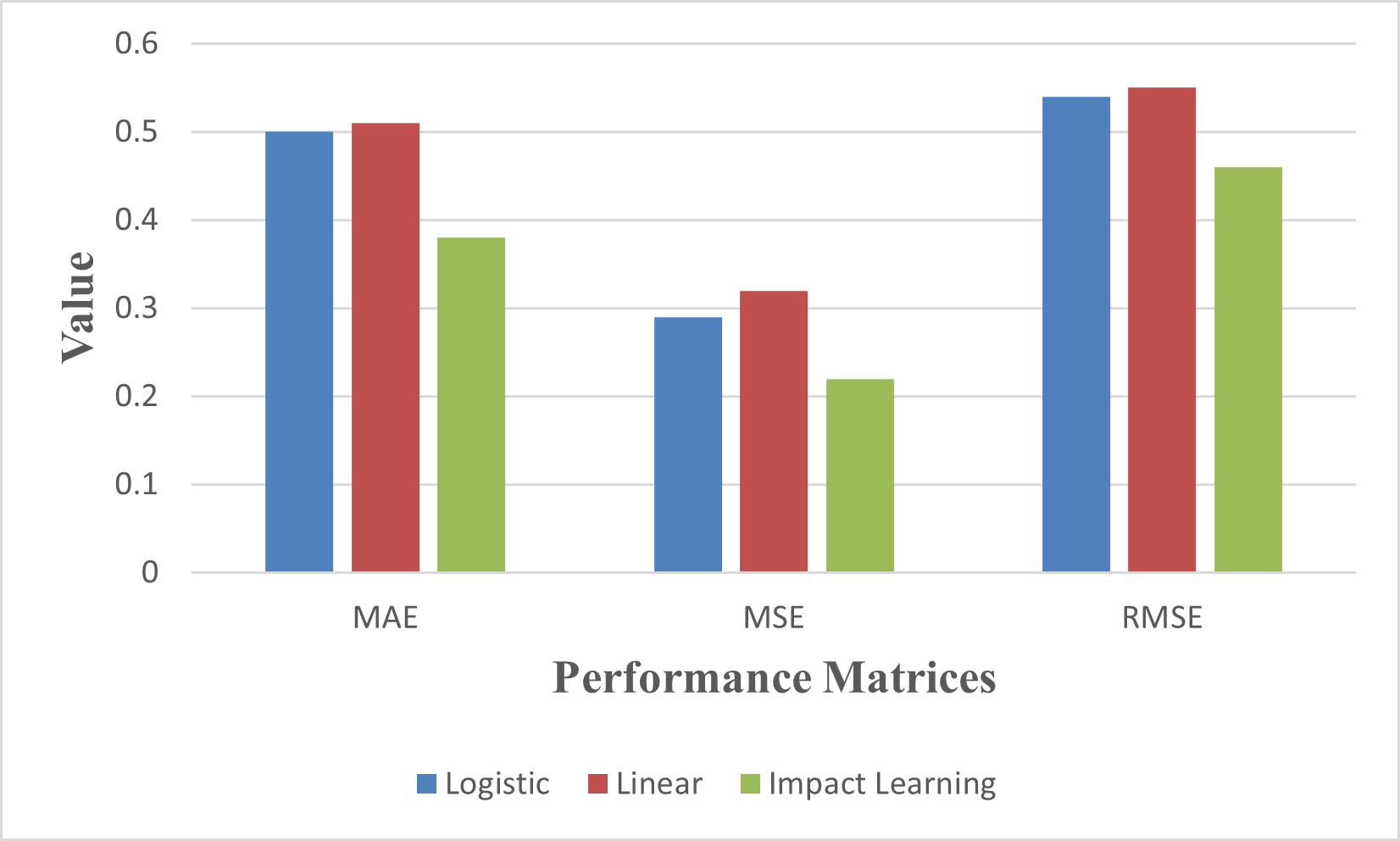}
    \caption{Performance analysis (MSE, MAE, and RMSE) for Logistic, Linear and Impact Learning algorithm.}
    \vspace{-5pt}
    \label{fig14}
\end{figure}

A heart illness dataset from Kaggle with 16 columns was used to improve the regression dataset's performance. The goal of this dataset was to forecast the likelihood of cardiovascular disease during the following ten years.We have considered three factors to analyze this dataset’s performance: the MSE, MAE, and RMSE depicted in Figure \ref{fig14}.

When it comes to improving the performance of our proposed algorithm (Impact Learning), we have benchmarked it against two of the most prominent algorithms available, namely linear regression and logistic regression. Figure \ref{fig14} indicates that the output of Impact Learning is superior to the output of two other algorithms for all of the performance matrices. Specifically, the effect learning value for MAE is less than 40\%, the MSE value is somewhat higher than 20\%, and the RMSE value is almost 50\%. We obtained the lowest performance with the linear approach, although it is practically identical to that obtained with logistic regression.

\begin{figure}[htb]
    \centering
    \includegraphics[width=0.9\linewidth]{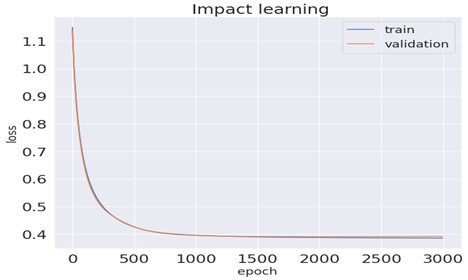}
    \caption{Training and validation loss curve for Impact Learning.}
    \vspace{-5pt}
    \label{fig15}
\end{figure}

Figure \ref{fig15} shows that the training and validation loss curves are shrinking as the number of epochs increases. Figure \ref{fig15} indicates that the performance of Impact learning is improving as the number of epochs increases.

\section{Conclusion and Future Work}
\label{sec:7}

We have developed a novel machine learning approach for handling regression and classification issues. The system of learning from RNI and the impact of additional aspects such as competition is the primary techniques employed by this strategy. As it gains knowledge from the effect, it can be applied to other aspects of competition, such as the effect of other actors. To improve the performance of this algorithm, we used three real datasets, two of which were used for classification analysis and one of which was used for regression analysis. We compared the performance of our method to that of several other well-known algorithms. When we reached the results of our proposed algorithm to those of different algorithms, we discovered that it produced an excellent outcome. Accuracy, precision, recall, and F1-score are some matrices that we examine while looking at performance evolution.

We hope to use impact learning to solve NLP challenges in the future, integrating it with existing machine learning and deep learning algorithms for improved performance. It will be trained using backpropagation and gradient descent rather than the least-squares approach. Additionally, we've developed a method for using this model to evaluate and forecast market value among many competitors in business, economics, and machine learning, among others. 
\bibliographystyle{IEEEtran}
 \bibliography{name}
\appendices

%--------------
% Appendix:02
%--------------
\vskip -2\baselineskip plus -1fil

\end{document}